\documentclass[10pt, conference, compsocconf]{IEEEtran}

\pdfpagewidth  = \paperwidth
\pdfpageheight = \paperheight

\usepackage{graphicx}
\usepackage{booktabs}
\usepackage{array}
\usepackage{cite}
\usepackage{url}
\usepackage{fixltx2e}

\usepackage{amssymb}
\usepackage{amsmath}
\usepackage[norelsize,ruled,linesnumbered]{algorithm2e} 
\usepackage[caption=false,font=footnotesize]{subfig}

\begin{document}

\title{Wikipedia Edit Number Prediction based on Temporal Dynamics Only}

\author{
	\IEEEauthorblockN{
		Dell Zhang
	}
	\IEEEauthorblockA{
		DCSIS\\
		Birkbeck, University of London\\
		Malet Street\\
		London WC1E 7HX, UK\\
		Email: dell.z@ieee.org
	}
}

\maketitle

\begin{abstract}
In this paper, we describe our approach to the Wikipedia Participation Challenge which aims to predict the number of edits a Wikipedia editor will make in the next 5 months. 
The best submission from our team, ``zeditor'', achieved 41.7\% improvement over WMF's baseline predictive model and the final rank of 3rd place among 96 teams. 
An interesting characteristic of our approach is that only temporal dynamics features (i.e., how the number of edits changes in recent periods, etc.) are used in a self-supervised learning framework, which makes it easy to be generalised to other application domains.
\end{abstract}


\section{Introduction}
\label{sec:Introduction}

Wikipedia is ``a free, web-based, collaborative, multilingual encyclopaedia project'' supported by the non-profit Wikimedia Foundation (WMF). Started in 2001, Wikipedia has become the largest and most popular general reference knowledge source on the Internet. Almost all of its 19.7 million articles can be edited by anyone with access to the site, and it has about 90,000 regularly active volunteer editors around the world. 
However, it has recently been observed that Wikipedia growth has slowed down significantly \cite{Suh2009}. In particular, WMF has reported that\footnote{\url{http://strategy.wikimedia.org/wiki/March_2011_Update}}: 
\begin{quote}
Between 2005 and 2007, newbies started having real trouble successfully joining the Wikimedia community. Before 2005 in the English Wikipedia, nearly 40\% of new editors would still be active a year after their first edit. After 2007, only about 12-15\% of new editors were still active a year after their first edit. Post-2007, lots of people were still trying to become Wikipedia editors. What had changed, though, is that they were increasingly failing to integrate into the Wikipedia community, and failing increasingly quickly. The Wikimedia community had become too hard to penetrate.
\end{quote}
It is therefore of utter importance to understand quantitatively what factors determine editors' future editing behaviour (why they continue editing, change the pace of editing, or stop editing), in order to ensure that the Wikipedia community can continue to grow in terms of size and diversity.

The Wikipedia Participation Challenge\footnote{\url{http://www.kaggle.com/c/wikichallenge}}, sponsored by WMF and hosted by Kaggle, request contestants to build a predictive model that could accurately predict the number of edits a Wikipedia editor would make in the next 5 months based on his edit history so far. Such a predictive model may be able to help WMF in figuring out how people can be encouraged to become, and remain, active contributors to Wikipedia.

The `training' dataset consists of randomly sampled active editors with their full history of editing activities on the English Wikipedia (the first 6 namespaces only) in the period from 2001-01-01 to 2010-09-01. An editor is considered ``active'' if he or she made at least one edit in the last one year period, i.e., from 2009-09-01 to 2010-09-01. For each edit, the available information includes its user\_id, article\_id, revision\_id, namespace, timestamp, etc.

The predictive model to be constructed should predict, for each editor from the `training' dataset, how many edits would be made in the 5 months after the end date of the `training' dataset, i.e., from 2010-09-01 to 2011-02-01. 
The predictive model's accuracy is going to be measured by the Root Mean Squared Logarithmic Error (RMSLE):
\begin{equation}
\epsilon = \sqrt{ \frac{1}{n} \sum_{i=1}^{n}{(\log(1+p_i)-\log(1+a_i))^2} } \ ,
\label{eq:RMSLE}
\end{equation}
where $n$ is total number of editors in the dataset, $\log(\cdot)$ is the natural logarithm function, $p_i$ and $a_i$ are the predicted and actual edit numbers respectively for editor $i$ in the next 5 month period.

The best submission from our team, ``zeditor'', achieved 41.7\% improvement over WMF's baseline predictive model and the final rank of 3rd place among 96 teams. 
An interesting characteristic of our approach is that only temporal dynamics features (i.e., how the number of edits changes in recent periods, etc.) are used in a self-supervised learning framework, which makes it easy to be generalised to other application domains.

The rest of this paper is organised as follows. 
In Section \ref{sec:Approach}, we present our approach in details.
In Section \ref{sec:Experiments}, we show the experimental results.
In Section \ref{sec:RelatedWork}, we review the related work.
In Section \ref{sec:Conclusions}, we make conclusions.

\section{Approach}
\label{sec:Approach}

\begin{figure*}[!t]
	\centering
		\includegraphics[width=0.6\textwidth]{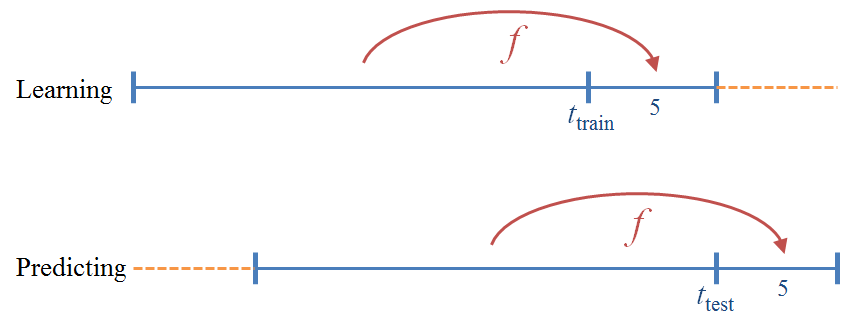}
	\caption{Our self-supervised learning framework.}
	\label{fig:framework}
\end{figure*}

Our basic idea is to build a predictive model $f$ (that estimates an active editor's future number of edits based on his recent edit history) through self-supervised learning, as illustrated schematically in Figure~\ref{fig:framework}.
The approach is called ``self-supervised'' to emphasise the fact that it does not require any manual labelling of data (as in standard \emph{supervised learning} \cite{Hastie2009}) but extracts the needed labels from data automatically.

To facilitate the description of our approach, we shall from now on talk about any time-length in the unit of months and refer to any time-point as the real number of months passed since the beginning date of the dataset. 
So for the official dataset `training', the timestamp ``2001-06-16 00:00:00'' would be 5.5 because it is five and a half months since 2001-01-01.

Let $t_\text{test}$ denote the time-point when we would like to predict each active editor's number of edits in the next 5 months. 
To train the predictive model, we would move 5 months backwards and assume that we were at the time-point $t_\text{train} = t_\text{test} - 5$. Thus we could know the actual number of edits made by each active editor in those 5 months after $t_\text{train}$, i.e., the label for our machine learning (regression) methods. 
Specifically, the target value for regression would be set as $y_i = \log(1+a_i)$ where $a_i$ is the actual number of edits in the next 5 months. In this way, the \emph{squared error} loss function $L(f(\mathbf{x}), y) = (f(\mathbf{x}) - y)^2$ used by most machine learning methods (including those in our experiments and final submission) would connect the \emph{empirical risk} \cite{Hastie2009} directly to the evaluation metric RMSLE: 
\begin{equation}
R_{emp}(f) = \frac{1}{n} \sum_{i=1}^n L(f(\mathbf{x}_i), y_i) = \epsilon^2 \ .
\label{eq:R_emp}
\end{equation}

Given a time-point (either $t_\text{train}$ or $t_\text{test}$), each active editor $i$ would be represented as a vector $\mathbf{x}_i$ that consists of the following temporal dynamics features:
\begin{itemize}
	\item the number of edits in recent periods of time; 
	\item the number of edited articles in recent periods of time;
	\item the length of time between the first edit and the last edit, scaled logarithmically. 
\end{itemize}
The periods used in our final submission for the above temporal dynamics features are 
\[ \frac{1}{16}, \frac{1}{8}, \frac{1}{4}, \frac{1}{2}, 1, 2, 4, 12, 36, 108 \] 
where the length of period first doubles at each step from $\frac{1}{16}$ to 4 and then triples at each step from 4 to 108. 
The usage of such temporal dynamics features was inspired by the decent performance of the ``most-recent-5-months-benchmark'' --- if using the exact number of edits in just one period (the last 5 months) for prediction could work reasonably well, we should be able to achieve a better performance by using many more recent periods.
The periods were chosen to be at exponentially increasing temporal scales, because we conjecture that the influence of an editing activity to the editor's future editing behaviour would be exponentially decaying along with the time distance away from now. 
The process of \emph{exponential decay}\footnote{\url{http://en.wikipedia.org/wiki/Exponential_decay}} occurs in numerous natural phenomena, and it has been widely used in temporal applications where it is desirable to gradually discount the history of past events \cite{Aggarwal2004}. 
One reason for changing from doubling to tripling midway through is to include the special period of 12-months (i.e., one year) that has been used to define the ``active'' editors.
The periods will be capped by the time scope of the given dataset (e.g., 106 for the additional dataset `moredata') in case they are out of range. 

We have also introduced a constant \emph{drift} term (i.e., how much the average number of edits would change after 5 months) into the formula of making final predictions, which is a crude way to cover the global shift of target values along with time. Again, its value is estimated from the situation 5 months ago. 

The concise pseudo-code of our algorithms for learning and predicting is shown in Figure~\ref{fig:algorithm}.
The complete source code will be made available at the author's homepage\footnote{\url{http://www.dcs.bbk.ac.uk/~dell/}}.

\begin{figure*}[!t]
	\rule[1mm]{\textwidth}{.8pt}
	Learning
	\begin{itemize}
		\item $t = t_\text{train}$ (i.e., $t_\text{test}-5$)
		\item for each active editor $i$ who made at least one edit in $[t-12, t)$:
		\begin{itemize}
			\item represent the editor as a vector $\mathbf{x}_i$ consisting of temporal dynamics features (please refer to the above description)
			\item label the editor by $y_i = \log(1+a_i)$ where $a_i$ is the actual number of edits in $[t, t+5)$
		\end{itemize}
		\item learn a predictive model/function $f: x \rightarrow y$ from $(\mathbf{x}_i,y_i)$ pairs using a regression technique such as GBT
		\item estimate the drift $d$ by comparing the average number of edits in $[t-5, t)$ and that in $[t, t+5)$
	\end{itemize}
	\rule[1mm]{\textwidth}{.5pt}
	Predicting
	\begin{itemize}
		\item $t = t_\text{test}$ (e.g., $116$ for the dataset `training')
		\item for each active editor $i$ who made at least one edit in $[t-12, t)$:
		\begin{itemize}
			\item represent the editor as a vector $\mathbf{x}_i$ consisting of temporal dynamics features (please refer to the above description)
			\item compute $\hat{y}_i$ = $f(\mathbf{x}_i)$ using the learnt $f$
			\item output $p_i = \exp(\max(\hat{y}_i+d,0))-1$ as the predicted number of edits in $[t, t+5)$
		\end{itemize}
	\end{itemize}
	\rule[1mm]{\textwidth}{.8pt}
	\caption{Our algorithms for learning and predicting.}
	\label{fig:algorithm}
\end{figure*}

\section{Experiments}
\label{sec:Experiments}

\subsection{Datasets}

There are three datasets available to all contestants: 
\begin{itemize}
	\item `training' is the official dataset for training and testing;
	\item `validation' is the official dataset for validation;
	\item `moredata' is the additional dataset generously provided by Twan van Laarhoven\footnote{\url{http://www.kaggle.com/c/wikichallenge/forums/t/719/more-training-data}}.
\end{itemize}
The characteristics of each dataset are shown in Table~\ref{tab:dataset}.

\begin{table}[!t]
	\renewcommand{\arraystretch}{1.3}
	\centering
	\caption{The characteristics of each dataset.}
	\label{tab:dataset}
	\begin{tabular}{|c|rrrr|}
	\hline  
	dataset & $t_\text{train}$ & $t_\text{test}$ & \#editors & \#edits \\
	\hline  
	`validation' & 79    & 84    & 4856   & 274820\\
	`moredata'   & 106   & 111   & 23584  & 5717049 \\
	`training'   & 111   & 116   & 44514  & 22326031 \\
	\hline  
	\end{tabular}
\end{table}

Since we did not have local access to the true labels (target values) of the dataset `training', we only used it to make the final submission, but conducted our experiments (for parameter tuning etc.) on the other two datasets `validation' and `moredata'.
It is noteworthy that these two datasets `validation' and `moredata' had been filtered to contain only active editors (who made at least one edit in the last one year period) in order to make them exhibit the same \emph{survivorship bias}\footnote{\url{http://en.wikipedia.org/wiki/Survivorship_bias}} as the dataset `training'. This might (partially) ensure that the experimental findings on the former two datasets could be transferred to the latter one.

\subsection{Tools}

We have only used Python\footnote{\url{http://www.python.org/}} (equipped with Numpy\footnote{\url{http://numpy.scipy.org/}}) to write small programs for analysing data and making predictions. 
The machine learning methods that we have tried for our regression task all come from two \emph{open-source} Python modules: one is scikit-learn\footnote{\url{http://scikit-learn.sourceforge.net/}}, and the other is OpenCV\footnote{\url{http://opencv.willowgarage.com/wiki/}}. 

\subsection{Results}

\begin{figure*}[!t]
	\centerline{
		\subfloat[validation]{
			\includegraphics[width=\columnwidth]{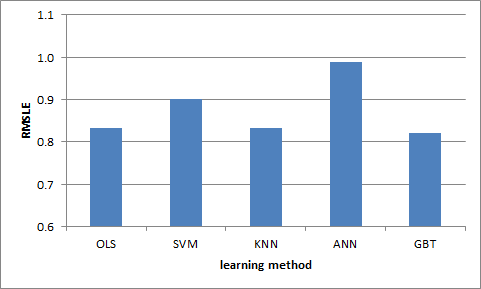}
			\label{fig:method_validation}
		}
		\hfil
		\subfloat[moredata]{
			\includegraphics[width=\columnwidth]{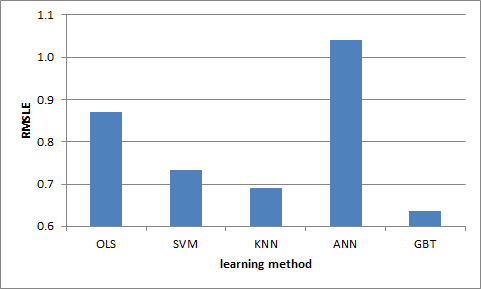}
			\label{fig:method_moredata}
		}
	}
	\caption{The prediction performances of different machine learning methods (with their default parameter values).}
	\label{fig:method}
\end{figure*}

First, we compare different machine learning methods (with their default parameter values) in terms of their prediction performances (RMSLE). 
The methods being compared include:
\begin{itemize}
	\item Ordinary Least Squares (OLS)\footnote{\url{http://scikit-learn.sourceforge.net/modules/linear_model.html#ordinary-least-squares-ols}},
	\item Support Vector Machine (SVM)\footnote{\url{http://scikit-learn.sourceforge.net/modules/svm.html}},
	\item K Nearest Neighbours (KNN)\footnote{\url{http://opencv.itseez.com/modules/ml/doc/k_nearest_neighbors.html}},
	\item Artificial Neural Network (ANN)\footnote{\url{http://opencv.itseez.com/modules/ml/doc/neural_networks.html}},
	\item Gradient Boosted Trees (GBT)\footnote{\url{http://opencv.itseez.com/modules/ml/doc/gradient_boosted_trees.html}}.
\end{itemize}
The experimental results are shown in Table~\ref{tab:method} and Figure~\ref{fig:method}. 
Gradient Boosted Trees (GBT)\footnote{\url{http://en.wikipedia.org/wiki/Gradient_boosting}} \cite{Friedman1999a,Friedman1999b} clearly outperformed all the other machine learning methods on both datasets.
GBT (aka GBM, MART and TreeNet) represents a general and powerful machine learning method that builds an ensemble of \emph{weak} tree learners in a greedy fashion. It evolved from the application of boosting to regression trees \cite{Hastie2009}. The general idea is to compute a sequence of very simple trees, where each successive tree is built for the prediction residuals of all preceding trees on a randomly selected subsample of the full training dataset. Eventually a ``weighted  additive expansion'' of those trees can produce an excellent fit of the predicted values to the observed values. It allows optimisation of any differentiable loss function. Here we just use the \emph{squared error} for the reasons given in Section \ref{sec:Approach}. 
The success of GBT in our task is probably attributable to (i) its ability to capture the complex nonlinear relationship between the target variable and the features, (ii) its insensitivity to different feature value ranges as well as outliers, and (iii) its resistance to overfitting via regularisation mechanisms such as shrinkage and subsampling \cite{Friedman1999a,Friedman1999b}. 


\begin{table}[!t]
	\renewcommand{\arraystretch}{1.3}
	\centering
	\caption{The prediction performances of different machine learning methods (with their default parameter values).}
	\label{tab:method}
	\begin{tabular}{|c|cc|}
	\hline  
	learning method & `validation' & `moredata' \\
	\hline  
	OLS   & 0.832351 & 0.869779 \\
	SVM   & 0.901698 & 0.732814 \\
	KNN   & 0.833288 & 0.690832 \\
	ANN   & 0.987345 & 1.040396 \\
	GBT   & 0.820805 & 0.635807 \\
	\hline  
	\end{tabular}
\end{table}

\begin{figure*}[!t]
	\centerline{
		\subfloat[validation]{
			\includegraphics[width=\columnwidth]{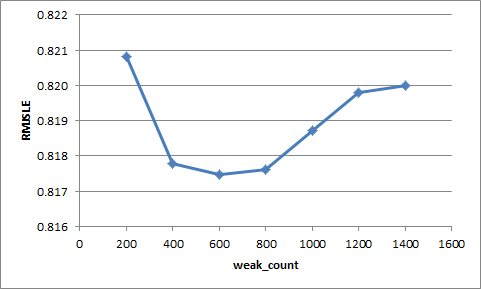}
			\label{fig:parameter_validation}
		}
		\hfil
		\subfloat[moredata]{
			\includegraphics[width=\columnwidth]{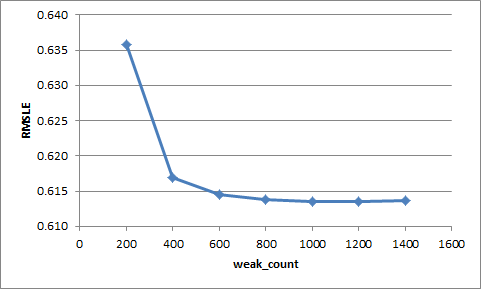}
			\label{fig:parameter_moredata}
		}
	}
	\caption{The prediction performances of GBT with different number of weak tree learners (weak\_count).}
	\label{fig:parameter}
\end{figure*}

Second, we investigate how GBT's most important parameter \textit{weak\_count} --- the number of weak tree learners --- affects its prediction performance for our task. Tuning \textit{weak\_count} is our major means of controlling the model complexity to avoid underfitting or overfitting.
The experimental results are shown in Table~\ref{tab:parameter} and Figure~\ref{fig:parameter}. 
It seems that on big datasets like `moredata', a higher value of \textit{weak\_count} (i.e., more weak tree learners) would be beneficial, but on small datasets like `validation', it might increase the risk of overfitting.

\begin{table}[!t]
	\renewcommand{\arraystretch}{1.3}
	\centering
	\caption{The prediction performances of GBT with different number of weak tree learners (weak\_count).}
	\label{tab:parameter}
	\begin{tabular}{|c|cc|}
	\hline  
	GBT \textit{weak\_count} & `validation' & `moredata' \\
	\hline  
	200   & 0.820805 & 0.635807 \\
	400   & 0.817789 & 0.616876 \\
	600   & 0.817483 & 0.614507 \\
	800   & 0.817614 & 0.613757 \\
	1000  & 0.818726 & 0.613530 \\
	1200  & 0.819804 & 0.613465 \\
	1400  & 0.819998 & 0.613671 \\
	\hline  
	\end{tabular}
\end{table}

\begin{figure*}[!t]
	\centerline{
		\subfloat[validation]{
			\includegraphics[width=\columnwidth]{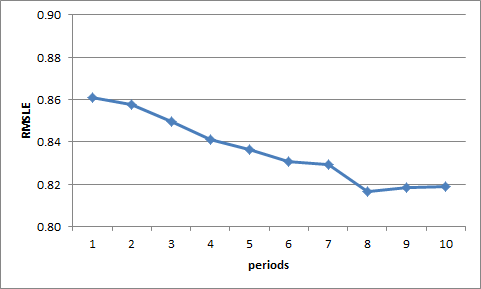}
			\label{fig:periods_validation}
		}
		\hfil
		\subfloat[moredata]{
			\includegraphics[width=\columnwidth]{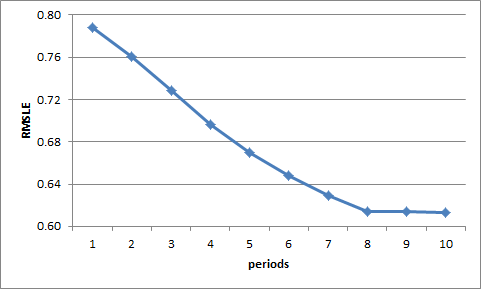}
			\label{fig:periods_moredata}
		}
	}
	\caption{The prediction performances of GBT using different number of periods for temporal dynamics features.}
	\label{fig:periods}
\end{figure*}

Third, we demonstrate how the prediction performance changes when we use more and more periods to generate temporal dynamics features: we start from just the shortest period ($\frac{1}{16}$) and then each time we add the next longer period to the series (see Section \ref{sec:Approach}). 
The experimental results are shown in Table~\ref{tab:periods} and Figure~\ref{fig:periods}. 
It seems that making use of more periods for temporal dynamics features usually helps, but the pay-off gradually diminishes. 

\begin{table}[!t]
	\renewcommand{\arraystretch}{1.3}
	\centering
	\caption{The prediction performances of GBT using different number of periods for temporal dynamics features.}
	\label{tab:periods}
	\begin{tabular}{|c|cc|}
	\hline  
	GBT \#periods & `validation' & `moredata' \\
	\hline  
	1  & 0.861111 & 0.788450 \\ 
	2  & 0.857575 & 0.760365 \\ 
	3  & 0.849440 & 0.728888 \\ 
	4  & 0.841127 & 0.696196 \\ 
	5  & 0.836116 & 0.669754 \\ 
	6  & 0.830619 & 0.647883 \\ 
	7  & 0.829393 & 0.629062 \\ 
	8  & 0.816459 & 0.614429 \\ 
	9  & 0.818515 & 0.613749 \\ 
	10 & 0.818726 & 0.613530 \\ 
	\hline  
	\end{tabular}
\end{table}

\subsection{Submissions}

Since `moredata' is more similar than `validation' to the official dataset `training' in terms of the time scope and the number of editors, we applied the best working algorithm, GBT, with the optimal parameter setting on `moredata' (\textit{weak\_count} = 1000), to make the final submission based on `training'. It got an RMSLE score of 0.862582 on the private leaderboard, which is roughly 41.7\% better than WMF's baseline predictive model. The final rank of our team, ``zeditor'', is the 3rd place among 96 teams.

\section{Related Work}
\label{sec:RelatedWork}

The global slowdown of Wikipedia's growth rate (both in the number of editors and the number of edits per month) has been studied \cite{Suh2009}. It is found that medium-frequency editors now cover a lower percentage of the total population while high frequency editors continue to increase the number of their edits. Moreover, there are increased patterns of conflict and dominance (e.g., greater resistance to new edits in particular those from occasional editors), which may be the consequence of the increasingly limited opportunities in making novel contributions. These findings could guide us to generate other kinds of useful features to tackle the problem of edit number prediction. Furthermore, researchers have also investigated other activities of Wikipedia's editors, such as voting on the promotion of Wikipedia admins \cite{Leskovec2010}.

In addition to Wikipedia, the temporal dynamics of online users' behaviour has been explored and exploited in web search \cite{Zhang2009b,Zhang2009a,Elsas2010,Kulkarni2011}, social tagging \cite{Zhang2009,Halpin2007}, blogging \cite{Lin2008}, twittering \cite{Abel2011}, and collaborative filtering \cite{Koren2009}. The \emph{power law} \cite{Clauset2009} and the \emph{exponential decay} \cite{Aggarwal2004} seem to be recurrent themes across application domains.

\section{Conclusions}
\label{sec:Conclusions}

Our most important insight is that a Wikipedia editor's future behaviour can be largely determined by the temporal dynamics of his recent behaviour. 
We are a bit surprised that just temporal dynamics features can go such a long way when we choose proper temporal scales and employ a powerful machine learning method. Human beings seem to be working and living in a more mechanical way than one might have thought.
Since such temporal dynamics features are actually independent of any semantics or knowledge about this specific problem, our approach could be easily generalised to other application domains, such as predicting the future supermarket spendings of shoppers (e.g., the dunnhumby's Shopper Challenge\footnote{\url{http://www.kaggle.com/c/dunnhumbychallenge}}), predicting the future hospital admissions of patients (e.g., the Heritage Health Prize Competition\footnote{\url{http://www.heritagehealthprize.com/c/hhp}}), and so on, based on historical behavioural data.

Have we answered the question that we asked at the beginning of this paper? Yes and No. 
On one hand, we have built a predictive model which can be used to identify those editors who are likely to become inactive, or in other words, who need special care and attention to be kept --- if an editor is going to leave the Wikipedia community, there would probably be early signals in the temporal dynamics of his recent behaviour.
On the other hand, that predictive model is pretty much a black box --- it does not reveal the underlying reasons why editors become inactive, and therefore it cannot tell us how to encourage editors to remain active. 
For the ultimate purpose of Wikipedia's sustainable growth, we will need to investigate which attributes of an editor (his articles' category distribution, his relationship with other editors, etc.) and also which recent events happened to him (his articles being deleted, his revisions being reverted, unfair comments about his edits being received, etc.) could affect his behaviour. 
Due to the time constraints and the dataset limitations (for example, the lack of information about articles and comments in the datasets `validation' and `moredata'), we have to leave it to future work.


Long live Wikipedia!

\section*{Acknowledgment}

We would like to thank WMF and Kaggle for their wonderful job in organising this interesting contest.
We are grateful to Twan van Laarhoven for creating and sharing the additional dataset `moredata'. 
We also appreciate the reviewers' helpful comments.

\bibliographystyle{IEEEtran}



\end{document}